\pdfoutput=1
\documentclass[11pt]{article}

\usepackage[]{acl}

\usepackage{times}
\usepackage{latexsym}

\usepackage[T1]{fontenc}

\usepackage[utf8]{inputenc}

\usepackage{color}
\usepackage{tabularx}
\usepackage{tabu}
\usepackage{booktabs}
\usepackage{amsmath}
\usepackage{multirow}

\usepackage[table]{xcolor}

\usepackage{graphicx} 
\usepackage{float}
\usepackage{subfigure} 
\usepackage{amssymb}
\usepackage{makecell}
\usepackage{array}
\usepackage{ragged2e}

\usepackage{hyperref}
\usepackage{cleveref}
\usepackage{tablefootnote}
\usepackage{xspace}

\usepackage{float}
\usepackage{amsmath}
\usepackage{bm}
\usepackage{algorithmicx}
\usepackage{algorithm}
\usepackage{algpseudocode}

\usepackage{arydshln}
\usepackage{gradient-text}

\usepackage{tcolorbox}
\usepackage{listings}
\tcbuselibrary{breakable, skins, listings}
\usepackage{caption}
\usepackage{cuted}
\usepackage{amssymb}
\usepackage{pifont}

\newcommand{\cmark}{{\color{green!40!black}\ding{51}}}
\newcommand{\xmark}{\ding{55}}%
\usepackage{enumitem}

\usepackage{placeins}

\newcommand{\huggingface}{\raisebox{-1.5pt}{\includegraphics[height=1.05em]{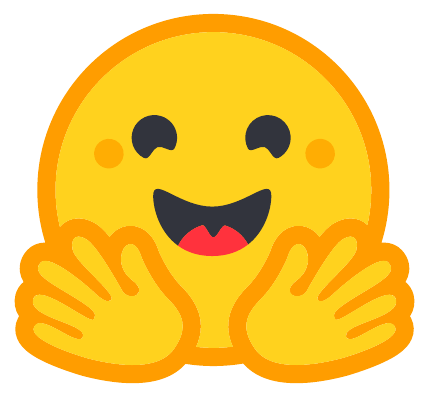}}\xspace}
\newcommand{\github}{\raisebox{-1.5pt}{\includegraphics[height=1.05em]{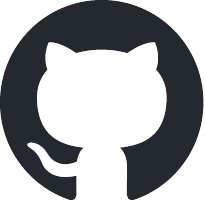}}\xspace}

\usepackage{microtype}

\newcommand{\ours}{TableVista\xspace}

\newcommand{\data}{TableVista\xspace}

\newcommand{\nmodel}{29\xspace}

\newcommand{\eg}{\hbox{\emph{e.g.,}}\xspace}
\newcommand{\ie}{\hbox{\emph{i.e.,}}\xspace}

\newcommand{\etc}{\hbox{\emph{etc.}}\xspace}

\title{\ours: Benchmarking Multimodal Table Reasoning under \\ Visual and Structural Complexity}

\author{
Zheyuan Yang$^1$ \quad  
Liqiang Shang$^2$ \quad 
Junjie Chen$^3$ \quad 
Xun Yang$^1$ \quad 
Chenglong Xu$^1$ \\
\bf{Bo Yuan$^1$ \quad 
Chenyuan Jiao$^1$ \quad 
Yaoru Sun$^{1\dagger}$ \quad 
Yilun Zhao$^4$} 
\vspace{3pt} \\
$^1$Tongji University \quad $^2$University of Bristol \quad $^3$Tianjin University \quad $^4$Yale University
}
\begin{document}

\maketitle
{\renewcommand{\thefootnote}{}\footnotetext{\raggedright $^\dagger$Correspondence to: Yaoru~Sun (\texttt{yaoru@tongji.edu.cn})}}

\begin{abstract}
We introduce \ours, a comprehensive benchmark for evaluating foundation models in multimodal table reasoning under visual and structural complexity. \ours consists of 3,000 high-quality table reasoning problems, where each instance is expanded into 10 distinct visual variants through our multi-style rendering and transformation pipeline. This process encompasses diverse scenario styles, robustness perturbations, and vision-only configurations, culminating in 30,000 multimodal samples for a multi-dimensional evaluation. We conduct an extensive evaluation of \nmodel state-of-the-art open-source and proprietary foundation models on \ours. Through comprehensive quantitative and qualitative analysis, we find that while evaluated models remain largely stable across diverse rendering styles,
they exhibit pronounced performance degradation on complex structural layouts and vision-only settings, revealing that current models struggle to maintain reasoning consistency when structural complexity combines with visually integrated presentations. These findings highlight critical gaps in current multimodal capabilities, providing insights for advancing more robust and reliable table understanding models.

\begin{small}
\begin{center}
\begin{tabular}{cll}
\huggingface & \textbf{Data} & \href{https://huggingface.co/TableVista} {\path{TableVista}}\\
\github & \textbf{Code} & \href{https://github.com/FlowRays/TableVista}{\path{FlowRays/TableVista}}\\
\end{tabular}
\end{center} 
\end{small}

\end{abstract}

\vspace{5pt}
\section{Introduction}

Tables serve as the bedrock of structured data across diverse professional domains, yet the optimal way for foundation models to perceive and reason over them remains an open question~\citep{deng-etal-2024-tables, zhou-etal-2025-texts, zhao-etal-2024-docmath}. Traditionally, table reasoning has relied on text-based serialization (\eg Markdown or HTML), but this approach often collapses the rich, spatial-structural information inherent in complex layouts~\citep{zhang-etal-2024-tablellama, yang-etal-2025-table}. In contrast, multimodal table reasoning (\ie treating tables as visual artifacts) naturally preserves intricate features such as multi-level headers, merged cells, and multi-table alignments that are frequently mangled in raw text strings~\citep{zheng-etal-2024-multimodal, NEURIPS2024_0d97fe65, Zhou_2025_CVPR}. This visual framing also better reflects real user interactions, in which tables are typically encountered as rendered documents or interface elements rather than as linearized data. However, the extent to which current multimodal foundation models can sustain consistent reasoning across diverse visual presentations remains underexplored, especially under structural complexity and real-world perturbations.

To bridge this gap and provide a rigorous assessment of multimodal table reasoning, we introduce \ours, a comprehensive benchmark designed to evaluate foundation models under conditions of high visual and structural complexity. Unlike existing datasets that often rely on simplistic or uniform table renderings, \ours comprises 3,000 high-quality reasoning problems, each expanded into 10 distinct visual variants through a sophisticated multi-style rendering and transformation pipeline. This pipeline systematically introduces diverse scenario styles, robustness perturbations, and real-world environmental perturbations, resulting in a large-scale collection of 30,000 multimodal samples. By providing such a multi-dimensional evaluation suite, \ours forces models to go beyond basic OCR-like recognition and demands a deeper synthesis of visual perception and logical reasoning across heterogeneous presentations.

\begin{table*}[t]
\centering
\resizebox{\textwidth}{!}{%
\begin{tabular}{l c ccc ccc}
\toprule
\multirow{2}{*}{\textbf{Benchmark}} & \multirow{2}{*}{\textbf{Multimodal}} & \multicolumn{3}{c}{\textbf{Structural Complexity}} & \multicolumn{3}{c}{\textbf{Visual Diversity \& Robustness}} \\
\cmidrule(lr){3-5} \cmidrule(lr){6-8}
 & & \textbf{Hierarch.} & \textbf{Long} & \textbf{Multi} & \textbf{Scenario} & \textbf{Transformation} & \textbf{Vision-Only} \\
\midrule
HiTab~\cite{cheng-etal-2022-hitab} & \xmark & \cmark & \xmark & \xmark & -- & -- & -- \\
NeedleInATable~\cite{wang2026needleinatable} & \xmark & \cmark & \cmark & \xmark & -- & -- & -- \\
MMQA~\cite{ICLR2025_794a425a} & \xmark & \xmark & \cmark & \cmark & -- & -- & -- \\
TableVQA-Bench~\cite{kim2024tablevqabenchvisualquestionanswering} & \cmark & \xmark & \xmark & \xmark & \xmark & \xmark & \xmark \\
MMTabQA~\cite{mathur-etal-2024-knowledge} & \cmark & \xmark & \xmark & \xmark & \xmark & \xmark & \xmark \\
MMTab~\cite{zheng-etal-2024-multimodal} & \cmark & \cmark & \cmark & \xmark & \xmark & \xmark & \xmark \\
MMTBench~\cite{titiya2025mmtbenchunifiedbenchmarkcomplex} & \cmark & \cmark & \cmark & \xmark & \xmark & \xmark & \xmark \\
\midrule
\textbf{\ours} (ours) & \textbf{\cmark} & \textbf{\cmark} & \textbf{\cmark} & \textbf{\cmark} & \textbf{\cmark} & \textbf{\cmark} & \textbf{\cmark} \\
\bottomrule
\end{tabular}%
}
\caption{Comparison of \textbf{\ours} with representative table understanding benchmarks. \ours is the first benchmark to integrate diverse structural complexities with a rigorous multi-style visual robustness pipeline.}
\label{tab:comparison}
\end{table*}

We conduct an extensive evaluation of \nmodel state-of-the-art foundation models, including both proprietary and open-source models. We find that performance degrades sharply as structural complexity and reasoning depth increase, with open-source models exhibiting more severe collapse. Beyond reasoning, visual conditions introduce an additional failure mode: while models remain stable across rendering themes, they are brittle under realistic perturbations such as fragmented layouts and phone-captured tables, where errors are driven more by spatial misalignment than text recognition. Prompting further reveals a stark contrast—chain-of-thought largely narrows the gap between open-source and proprietary models, yet substantial disparities re-emerge under direct-output prompting, indicating limited internalization of multi-step tabular reasoning. These findings highlight critical bottlenecks in current multimodal models, particularly in their ability to robustly anchor structured reasoning in visual space.

\section{Related Work}
\paragraph{Table Reasoning Evaluation.}
Early evaluation benchmarks for table reasoning primarily focused on text-only serialization, addressing tasks like question answering~\cite{pasupat-liang-2015-compositional, nan-etal-2022-fetaqa, zhao-etal-2023-qtsumm, zhao-etal-2023-robut} and fact verification~\cite{chen2020tabfactlargescaledatasettablebased}.
However, these unimodal benchmarks overlook the rich spatial structure inherent in visual layouts. 
Recent benchmarks have begun to incorporate visual modalities. For instance, TableVQA-Bench~\cite{kim2024tablevqabenchvisualquestionanswering} and MMTab~\cite{zheng-etal-2024-multimodal} utilize synthetic rendering pipelines to evaluate visual reasoning capabilities. 
To account for richer table content and visual elements, MMTabQA~\cite{mathur-etal-2024-knowledge} introduces knowledge-aware reasoning over tables containing symbols and icons, while MMTBench~\cite{titiya2025mmtbenchunifiedbenchmarkcomplex} incorporates interleaved charts and images. In specific domains, MMSci~\cite{yang2025doestablesourcematter} targets numerical reasoning within scientific tables. Most recently, TABLET~\cite{alonso2026tablet} has emphasized large-scale robust evaluation on tables rendered from original web sources. 
However, as shown in \autoref{tab:comparison}, existing benchmarks typically present tables in a fixed, ideal, or single original visual state. They neglect the critical dimension of reasoning consistency under visual perturbations, where a model should yield identical answers despite changes in layout, style, or orientation. 

\paragraph{Multimodal Table Reasoning.}
Recent progress in multimodal table understanding has been driven by MLLMs, which enable direct reasoning over rendered tables and other visually grounded document elements.
Pioneering generalist models such as Table-LLaVA~\cite{zheng-etal-2024-multimodal} and TabPedia~\cite{NEURIPS2024_0d97fe65} enhance performance through table-specific instruction tuning and concept synergy mechanisms.
To address specific limitations in data and resolution, SynTab-LLaVA~\cite{Zhou_2025_CVPR} introduces a decoupled synthesis pipeline to scale up training data efficiently, while MMSci~\cite{yang2025doestablesourcematter} employs dynamic input resolution to capture fine-grained details in scientific tables.
More recently, research has pivoted toward advanced architectures and reasoning strategies. TableMoE~\cite{zhang2025tablemoeneurosymbolicroutingstructured} proposes a neuro-symbolic routing mechanism within a Mixture-of-Experts (MoE) framework to handle structurally degraded ``WildStruct'' tables. To bridge the modality gap, Turbo~\cite{jiang2026multimodal} leverages privileged structured information during training to distill reasoning capabilities from strong reasoning models such as DeepSeek-R1~\cite{Guo_2025}.

\begin{figure*}[!t]
 \centering
\includegraphics[width=\textwidth]{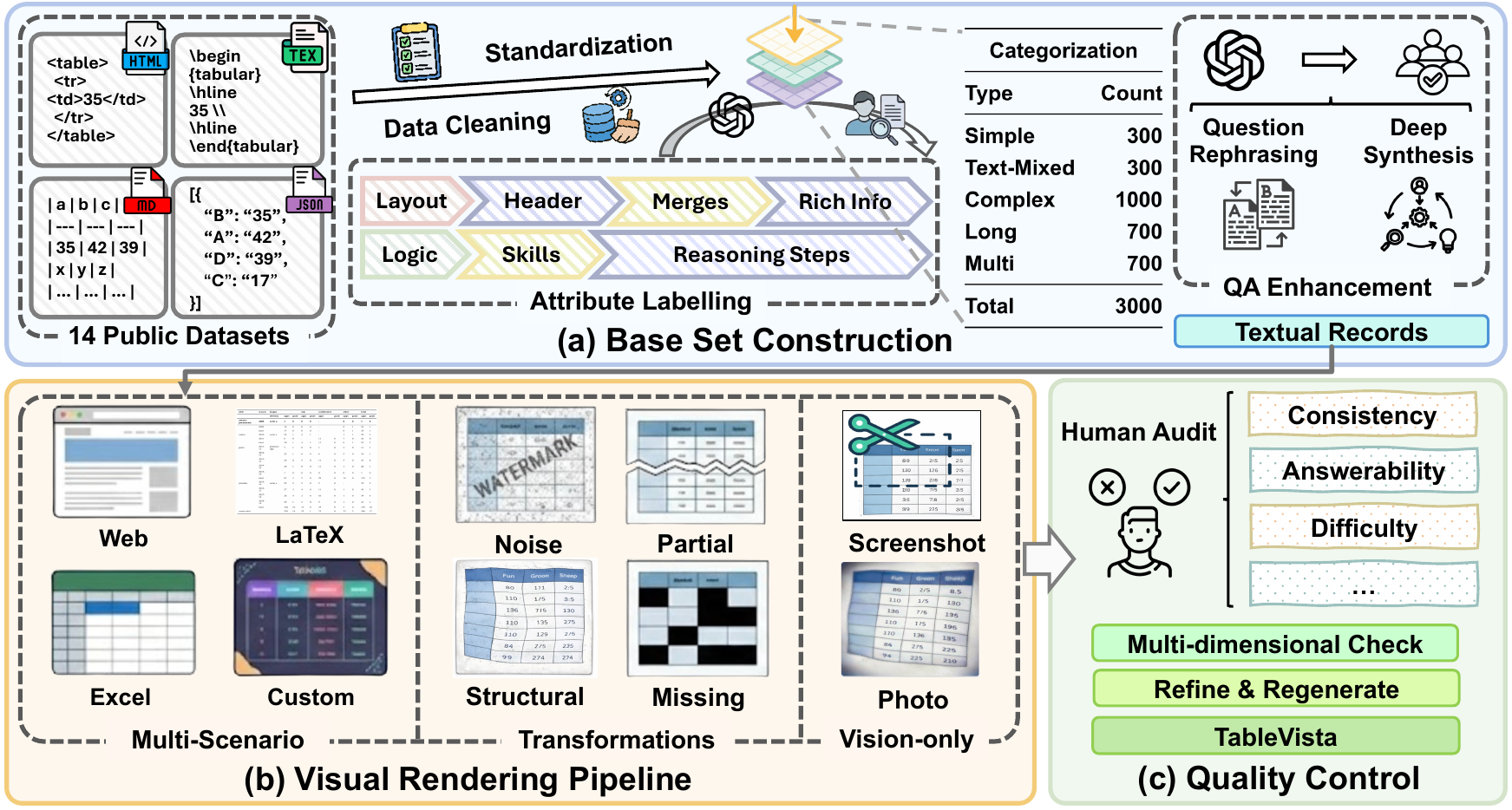}
 \caption{An overview of the \ours benchmark construction pipeline.}
 \label{fig:data_construction_pipeline}
\end{figure*}

\section{\data Evaluation Benchmark}
\label{sec:3}
This section details the construction process of the \ours benchmark. The pipeline is designed to establish a rigorous benchmark for evaluating the table reasoning capabilities of MLLMs by integrating diverse table structural complexities with challenging visual transformations. \autoref{fig:data_construction_pipeline} provides an overview of benchmark construction workflow.

\subsection{Base Set Construction}
\label{sec:3.1}
We first construct a textual \textbf{Base Set} by collecting and re-annotating samples from public table reasoning datasets. We engage 12 expert annotators (detailed information provided in Appendix \ref{app:annotators}) to re-annotate these samples with the assistance of state-of-the-art LLMs\footnote{Unless otherwise specified, we use GPT-5 by default throughout the data construction process.}. The Base Set construction process is as follows:

\begin{figure*}[!t]
 \centering
 \includegraphics[width=\textwidth]{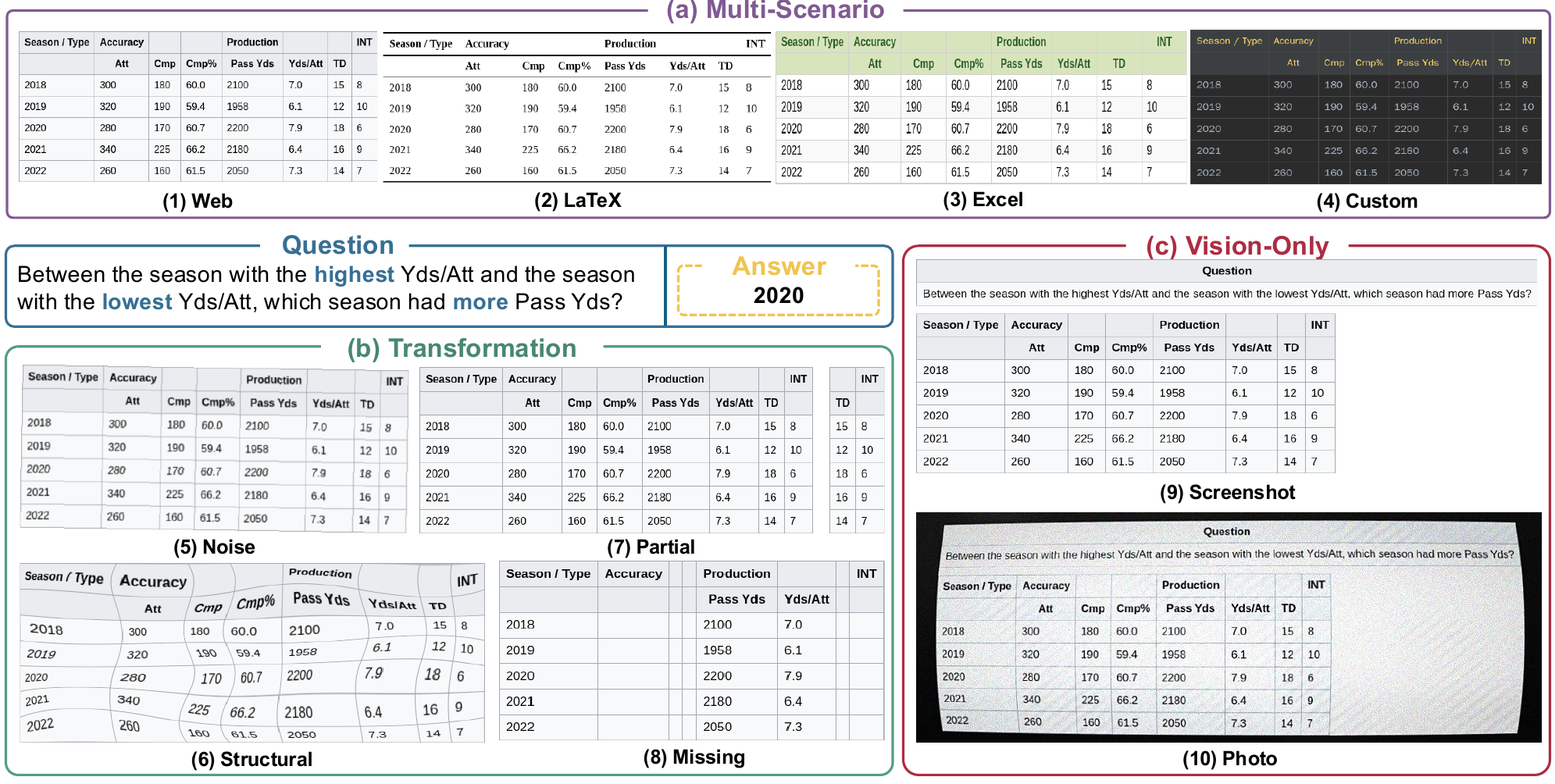}
 \caption{Samples of \ours. The central box illustrates a standard question–answer pair requiring calculation and reasoning over the table. (a) Multi-Scenario: four rendering styles—Web, LaTeX, Excel, and Custom. (b) Transformation: four robustness perturbations—Noise, Structural, Partial, and Missing. (c) Vision-Only: two settings simulating real-world capture—Screenshot and Photo.}
 \label{fig:10_samples}
\end{figure*}
\paragraph{Data Standardization and Attribute Labeling.}
We first aggregate raw records from 14 prominent public datasets (\eg WTQ~\cite{pasupat-liang-2015-compositional}, HiTab~\cite{cheng-etal-2022-hitab}, TabFact~\cite{chen2020tabfactlargescaledatasettablebased}, MMQA~\cite{ICLR2025_794a425a}, FinQA~\cite{chen-etal-2021-finqa}, \etc), standardizing them into a canonical format comprising the \textit{table}, \textit{textual context}, \textit{question}, and \textit{answer}.
For each record, we perform structured attribute labeling consisting of four tasks: (1) identifying table layout attributes (\eg multi-level headers, merged cells, composite structures) and recording dimensional information, (2) assigning an \textit{information richness} score (1--5) based on data diversity, relation density, and numerical computability, (3) characterizing the reasoning requirements of each QA pair by assigning \textit{skill scores} across four dimensions---lookup, aggregation, numerical, and logical---and (4) counting the total \textit{reasoning steps} required to derive the answer.
These labels are initially produced by GPT-5, and the annotators then verify each result by inspecting the model's chain-of-thought, re-annotating any cases they find incorrect.

\paragraph{Taxonomic Categorization and Multidimensional Filtering.}
Based on the labeled attributes, we perform rigorous filtering across two dimensions: table structural complexity and QA reasoning difficulty. First, we categorize records into five structural types with explicit quotas: \textit{Simple Structure} (standard grid), \textit{Text-Mixed} (with associated textual context), \textit{Complex Structure} (with hierarchical headers or merged cells), \textit{Long Tables} (with large row or column dimensions), and \textit{Multi-Table} scenarios (involving 2--5 associated tables). Within each structural quota, we prioritize samples with the highest reasoning potential. The selection criteria emphasize higher table \textit{information richness} scores, higher aggregate \textit{skill scores}, and larger numbers of \textit{reasoning steps}. Through this hierarchical selection procedure, we ultimately retain 3,000 highly representative and reasoning-intensive records, yielding a benchmark that balances structural diversity and reasoning capability.

\paragraph{QA Enhancement.}
For the filtered samples, annotators further enhance QA pairs using a dual-track strategy: (1) \textit{Question Rephrasing}: if the original QA already meets the target reasoning complexity, annotators rewrite the question to increase linguistic diversity and reduce potential contamination while preserving the answer; and (2) \textit{Deep Synthesis}: if the original QA is trivial, annotators author a new multi-step reasoning question grounded in the table and textual context, requiring at least two core skills and multiple logical hops. To ensure evaluative precision, all answers are standardized as unique short phrases or numbers.

\subsection{Multi-Style Visual Pipeline}
\label{sec:3.2}
The primary contribution of \ours is its sophisticated visual rendering pipeline, which systematically maps textual foundations onto a broad spectrum of visual representations, spanning diverse stylistic scenarios, structural and environmental perturbations, and unified vision-only configurations. This multi-faceted framework enables a rigorous evaluation of model robustness by simulating the complex interplay between structural layout, aesthetic variation, and simulated visual degradation, as illustrated in \autoref{fig:10_samples}.

\paragraph{Multi-Scenario Rendering.} 
To capture the diversity of table formats encountered in practice, we simulate four authentic scenarios that reflect distinct data ecosystems. The \textit{Web} scenario replicates the standard HTML layouts common in Wikipedia, while \textit{LaTeX} mimics the professional typography found in academic literature. The \textit{Excel} scenario reproduces the dense, grid-centric interfaces characteristic of spreadsheet applications. Finally, the \textit{Customized} scenario introduces high-variance visual noise by sampling from ten predefined thematic styles featuring diverse font families and color palettes. All scenarios are instantiated via a headless browser where viewports are dynamically optimized using localized bounding box metadata. This ensures that the resulting high-resolution captures maintain structural integrity without unintended cropping.

\paragraph{Robustness Transformations.}
Beyond stylistic variation, we introduce a suite of transformations to simulate the degraded or complex conditions often found in ``in-the-wild'' capture environments. Using the \textit{Web} scenario as a baseline, we apply four distinct perturbations. \textit{Noise} transformations incorporate stochastic combinations of Gaussian speckle, resolution downsampling, rotational tilt, and translucent watermarks. \textit{Structural Noise} transformations introduce structure-specific degradations, such as faded gridlines and geometric warping, that disrupt the table's visual skeleton while preserving its textual content. The \textit{Partial} transformation segments the table into disjoint visual blocks along structural boundaries to test long-range spatial reasoning. Lastly, \textit{Missing} transformations mask arbitrary cell contents without row- or column-level restrictions, while annotators verify that the remaining cells retain sufficient information to answer the target query.

\paragraph{Vision-Only Setting and Realism Artifacts.}
To align the evaluation with naturalistic user interactions, we propose a \textit{Vision-Only} configuration where the model processes a single image without separate text input. This setting is operationalized through two distinct modalities. The \textit{Screen Capture} mode renders the question, context, and table into a single unified interface, while the \textit{Simulated Photo} mode applies synthetic camera artifacts, such as moiré patterns and perspective distortion, directly on top of the \textit{Screen Capture} images to emulate the visual degradation of mobile photography. These challenges necessitate a synergistic capability to extract questions and interpret tabular data within a purely visual domain.

\begin{table}[t]
\centering
\small
\setlength{\tabcolsep}{12pt} 
\begin{tabular}{lr}
\toprule
\textbf{Metric} & \textbf{Count / Value} \\
\midrule
\multicolumn{2}{c}{\textbf{\textit{Dataset Size}}} \\[3pt]
Tables & 4,449 \\
QA Pairs & 3,000 \\
Visual Samples & 30,000 \\
\midrule
\multicolumn{2}{c}{\textbf{\textit{Structural Distribution}}} \\[3pt]
Simple Structure & 300 \\
Text-Mixed & 300 \\
Complex Structure & 1,000 \\
Long Table & 700 \\
Multi-Table & 700 \\
\midrule
\multicolumn{2}{c}{\textbf{\textit{Content Statistics}}} \\[3pt]
Avg. Question Words & 26.2 \\
Avg. Answer Words & 1.4 \\
Avg. Table Rows & 15.3 \\
Avg. Table Columns & 6.6 \\
\midrule
\multicolumn{2}{c}{\textbf{\textit{Reasoning Complexity (Steps / Skill Score)}}} \\[3pt]
Easy & 3.7 / 11.2 \\
Medium & 5.2 / 13.8 \\
Hard & 6.9 / 15.1 \\
\bottomrule
\end{tabular}
\caption{\ours data statistics.}
\end{table}

\subsection{Quality Control and Human Audit}
\label{sec:3.3}
We conduct a \emph{human-in-the-loop} audit throughout the pipeline, covering both the 3{,}000 textual base samples and the 30{,}000 rendered images produced by the visual pipeline. Expert reviewers perform a multi-dimensional check across consistency, answerability, difficulty, and other quality attributes, with particular attention to ensuring answerability even after severe visual transformations such as masking or splitting. Samples that fail any dimension are refined or regenerated until they meet the required standard. The resulting benchmark, \ours, demonstrates high logical rigor and scenario diversity, providing a superior quality benchmark for multimodal table understanding. Detailed rejection and revision statistics at each stage of the pipeline are reported in Appendix~\ref{app:quality_control}.

\begin{table*}[!t]
\centering
\setlength{\tabcolsep}{4pt}
\renewcommand\arraystretch{1.05}
\small
\footnotesize
\begin{tabular}{l ccccc ccc c}
\toprule[1pt]
\multirow{2}{*}{\textbf{Model}} & \multicolumn{5}{c}{\textbf{Structural Type}} & \multicolumn{3}{c}{\textbf{Reasoning Difficulty}} & \multirow{2}{*}{\textbf{Overall}} \\
\cmidrule(lr){2-6} \cmidrule(lr){7-9}
& \textbf{Simple} & \textbf{Text-Mixed} & \textbf{Complex} & \textbf{Long} & \textbf{Multi} & \textbf{Easy} & \textbf{Medium} & \textbf{Hard} & \\
\midrule[0.7pt]
\multicolumn{10}{c}{\textit{Proprietary Models}} \\
GPT-5.4 & \textbf{73.0} & \textbf{86.7} & \textbf{81.7} & \textbf{68.9} & \textbf{61.3} & \textbf{93.6} & \textbf{80.1} & \textbf{47.0} & \textbf{73.6} \\
GPT-5.4-mini & 52.0 & \underline{61.0} & 59.3 & 48.7 & 40.0 & 64.0 & \underline{56.1} & \underline{35.2} & 51.8 \\
\midrule
\multicolumn{10}{c}{\textit{Open-Source Models}} \\
Qwen2.5-VL-72B & 52.3 & 58.3 & \underline{59.7} & 52.1 & \underline{53.1} & \underline{90.2} & 54.3 & 22.1 & \underline{55.5} \\
Gemma-4-31B-it & \underline{57.3} & 54.0 & 57.6 & \underline{54.4} & 52.3 & 88.2 & 55.6 & 21.9 & 55.2 \\
Llama-4-Maverick & 55.3 & 55.7 & 55.9 & 52.3 & 52.4 & 84.4 & 53.7 & 24.4 & 54.2 \\
Qwen2.5-VL-32B & 51.0 & 49.7 & 54.5 & 48.3 & 49.1 & 84.5 & 49.0 & 19.4 & 51.0 \\
Qwen3.5-27B & 50.3 & 45.3 & 53.0 & 50.9 & 48.6 & 84.2 & 47.5 & 19.6 & 50.4 \\
Llama-4-Scout & 50.7 & 47.7 & 49.6 & 49.7 & 48.1 & 81.8 & 47.0 & 18.8 & 49.2 \\
Qwen3.5-122B-A10B & 45.0 & 48.3 & 52.2 & 49.4 & 46.3 & 82.6 & 45.7 & 18.9 & 49.1 \\
InternVL3.5-14B & 44.3 & 48.0 & 48.9 & 46.4 & 41.7 & 83.0 & 42.1 & 13.2 & 46.1 \\
Gemma-4-26B-A4B-it & 45.3 & 45.7 & 50.1 & 44.1 & 42.0 & 79.7 & 41.4 & 16.6 & 45.9 \\
InternVL3.5-30B-A3B & 46.7 & 49.0 & 47.7 & 43.3 & 43.3 & 81.6 & 41.3 & 14.1 & 45.7 \\
Qwen3-VL-30B-A3B & 44.3 & 48.7 & 46.8 & 42.4 & 41.0 & 78.9 & 39.4 & 14.8 & 44.4 \\
Qwen2.5-VL-7B & 43.0 & 41.3 & 47.5 & 41.9 & 42.4 & 79.0 & 38.6 & 14.2 & 43.9 \\
Gemma-3-27B-it & 46.7 & 44.0 & 44.6 & 43.7 & 39.6 & 76.5 & 39.5 & 14.1 & 43.4 \\
Qwen3-VL-8B & 40.7 & 44.0 & 44.1 & 41.9 & 39.9 & 76.7 & 37.3 & 12.7 & 42.2 \\
Molmo2-8B & 39.0 & 40.7 & 45.2 & 38.9 & 39.4 & 75.9 & 36.0 & 12.0 & 41.3 \\
Qwen3.6-35B-A3B & 41.3 & 38.0 & 42.5 & 40.3 & 41.0 & 70.9 & 35.9 & 16.4 & 41.1 \\
Qwen3.5-35B-A3B & 44.0 & 38.0 & 42.9 & 38.6 & 38.3 & 70.6 & 37.1 & 13.6 & 40.4 \\
Qwen3.5-9B & 42.3 & 41.3 & 40.3 & 39.1 & 39.4 & 72.6 & 33.7 & 14.1 & 40.1 \\
InternVL3.5-8B & 35.7 & 43.0 & 40.4 & 38.4 & 39.1 & 73.9 & 34.5 & 9.9 & 39.4 \\
Gemma-3-12B-it & 39.7 & 37.3 & 42.0 & 36.7 & 34.4 & 70.6 & 32.8 & 11.5 & 38.3 \\
MiniCPM-V-4.5 & 33.3 & 34.3 & 37.4 & 33.3 & 33.1 & 63.5 & 28.3 & 12.4 & 34.7 \\
Gemma-4-E4B-it & 30.3 & 23.7 & 31.0 & 26.3 & 26.1 & 53.1 & 23.2 & 7.6 & 28.0 \\
LLaVA-v1.6-Vicuna-13B & 16.0 & 17.0 & 18.6 & 15.1 & 14.9 & 30.8 & 12.8 & 5.9 & 16.5 \\
LLaVA-v1.6-Vicuna-7B & 14.0 & 16.3 & 16.9 & 13.4 & 14.6 & 27.1 & 13.2 & 5.3 & 15.2 \\
Table-LLaVA-v1.5-13B & 13.3 & 11.0 & 11.9 & 12.7 & 11.7 & 20.6 & 10.4 & 5.3 & 12.1 \\
Table-LLaVA-v1.5-7B & 11.0 & 11.0 & 7.8 & 9.4 & 9.3 & 16.7 & 6.8 & 4.0 & 9.2 \\
LLaVA-v1.5-7B & 6.0 & 5.7 & 6.6 & 6.4 & 6.7 & 10.6 & 4.7 & 4.0 & 6.4 \\
\bottomrule[1pt]
\end{tabular}
\caption{Model performance in the \textit{Web} scenario, stratified by structural complexity and reasoning difficulty.}
\label{tab:complexity_results}
\end{table*}
\section{Experiments}
In this section, we first describe the experimental setup and then provide a detailed analysis of the experimental results.

\subsection{Experimental Setup}
\paragraph{Evaluated Models.}
We evaluate the performance of state-of-the-art foundation models on our benchmark, including both proprietary and open-source models spanning a wide range of scales and reasoning capabilities. Proprietary models evaluated are GPT-5.4~\cite{openai_gpt5-4} and GPT-5.4-mini~\cite{openai_gpt5-4-mini}. Open-source models include Qwen2.5-VL~\cite{bai2025qwen25vltechnicalreport}, Qwen3-VL~\cite{bai2025qwen3vltechnicalreport}, Qwen3.5~\cite{qwen35blog}, Qwen3.6~\cite{qwen36_35b_a3b}, InternVL3.5~\cite{wang2025internvl35advancingopensourcemultimodal}, Llama-4~\cite{llama-4}, Gemma-3~\cite{gemmateam2025gemma3technicalreport}, Gemma-4~\cite{gemma-4}, Molmo2-8B~\cite{molmo2technicalreport}, MiniCPM-V-4.5~\cite{yu2025minicpmv45cookingefficient}, LLaVA-v1.5~\cite{Liu_2024_CVPR}, LLaVA-v1.6~\cite{liu2024llavanext}, and Table-LLaVA~\cite{zheng-etal-2024-multimodal}. Detailed model specifications and inference configurations are provided in Appendix~\ref{app:model_info}.

\paragraph{Evaluation Metrics.}
We employ a hybrid evaluation strategy combining Exact Match (EM) and an LLM-based Judge. We primarily assess correctness via normalized string matching. To mitigate penalties from minor formatting deviations, we utilize GPT-5-mini to verify semantic consistency for samples failing the EM criterion.

\paragraph{Evaluation Prompts.}
\label{sec:exp_setup}
For the analyses in \S\ref{sec:exp_complexity} and \S\ref{sec:exp_robustness}, we evaluate all models under a \textit{direct-output} prompt without enabling thinking mode (for GPT-5.4 and GPT-5.4-mini, we correspondingly set \texttt{reasoning\_effort} to \texttt{none}). We adopt this protocol because our preliminary experiments reveal that explicit reasoning compresses the performance landscape: under a CoT prompt with thinking mode enabled, even Qwen3-VL-8B-Thinking reaches 86.0\% on \ours, narrowing its gap with both proprietary SOTA and substantially larger open-source SOTA models to a few percentage points. In contrast, under the direct-output setting Qwen3-VL-8B-Instruct attains only 43.8\%, and pronounced gaps re-emerge across proprietary SOTA, open-source SOTA, and smaller open-source models. We therefore first analyze model performance along structural, reasoning, and visual-robustness dimensions under the direct-output setting, and subsequently provide a dedicated analysis of CoT-versus-direct-output behavior in \S\ref{sec:cot_vs_do}. The full prompt templates for both settings are provided in Appendix~\ref{app:prompt}.

\subsection{Evaluation on Table Structural and Reasoning Complexity}
\label{sec:exp_complexity}

\autoref{tab:complexity_results} reports the performance breakdown across five structural categories and three reasoning difficulty levels. Our primary findings are as follows:

\begin{table*}[!t]
\centering
\setlength{\tabcolsep}{4pt}
\renewcommand\arraystretch{1.05}
\resizebox{\textwidth}{!}{%
\footnotesize
\begin{tabular}{l cccc cccc cc c}
\toprule[1pt]
\multirow{2}{*}{\textbf{Model}} & \multicolumn{4}{c}{\textbf{Scenario Style}} & \multicolumn{4}{c}{\textbf{Visual Transformation}} & \multicolumn{2}{c}{\textbf{Vision-Only}} & \multirow{2}{*}{\textbf{Avg.}} \\
\cmidrule(lr){2-5} \cmidrule(lr){6-9} \cmidrule(lr){10-11}
& \textbf{Web} & \textbf{LaTeX} & \textbf{Excel} & \textbf{Custom} & \textbf{Noise} & \textbf{Structural} & \textbf{Partial} & \textbf{Missing} & \textbf{Screenshot} & \textbf{Photo} & \\
\midrule[0.7pt]
\multicolumn{12}{c}{\textit{Proprietary Models}} \\
GPT-5.4 & \textbf{73.6} & \textbf{72.2} & \textbf{71.9} & \textbf{72.0} & \textbf{70.8} & \textbf{70.4} & \textbf{68.8} & \textbf{84.8} & \textbf{69.4} & \textbf{67.3} & \textbf{72.1} \\
GPT-5.4-mini & 51.8 & 49.9 & 50.1 & 51.1 & 49.5 & 48.4 & 46.8 & 66.9 & 42.0 & 37.7 & 49.4 \\
\midrule
\multicolumn{12}{c}{\textit{Open-Source Models}} \\
Qwen2.5-VL-72B & \underline{55.5} & 54.5 & 54.5 & \underline{55.0} & 51.1 & \underline{54.4} & 50.5 & 71.3 & \underline{57.4} & \underline{54.0} & \underline{55.8} \\
Llama-4-Maverick & 54.2 & 53.2 & 53.9 & 52.7 & 53.5 & \underline{54.4} & \underline{52.9} & 66.9 & 53.4 & 51.8 & 54.7 \\
Gemma-4-31B-it & 55.2 & \underline{56.5} & \underline{55.9} & 54.4 & \underline{54.3} & 53.3 & 51.2 & \underline{75.7} & 42.2 & 44.2 & 54.3 \\
Qwen2.5-VL-32B & 51.0 & 50.7 & 50.7 & 51.9 & 49.3 & 49.1 & 48.1 & 67.3 & 51.1 & 49.8 & 51.9 \\
Qwen3.5-27B & 50.4 & 49.4 & 48.3 & 50.2 & 49.2 & 47.3 & 46.7 & 63.8 & 55.1 & 53.5 & 51.4 \\
Qwen3.5-122B-A10B & 49.1 & 48.5 & 48.6 & 48.9 & 49.0 & 48.7 & 47.3 & 63.3 & 54.3 & 53.5 & 51.1 \\
Llama-4-Scout & 49.2 & 48.6 & 47.7 & 48.1 & 47.5 & 48.3 & 46.0 & 64.9 & 46.7 & 47.2 & 49.4 \\
InternVL3.5-14B & 46.1 & 45.5 & 45.1 & 45.4 & 45.3 & 45.2 & 43.2 & 63.0 & 42.7 & 41.6 & 46.3 \\
InternVL3.5-30B-A3B & 45.7 & 44.4 & 44.4 & 45.6 & 45.6 & 44.9 & 43.5 & 65.1 & 40.1 & 35.5 & 45.5 \\
Qwen2.5-VL-7B & 43.9 & 44.3 & 44.4 & 44.1 & 42.5 & 42.1 & 41.0 & 61.8 & 41.6 & 40.0 & 44.6 \\
Qwen3-VL-30B-A3B & 44.4 & 42.7 & 42.7 & 45.4 & 43.0 & 41.8 & 40.4 & 60.2 & 41.1 & 42.4 & 44.4 \\
Qwen3-VL-8B & 42.2 & 41.8 & 41.7 & 42.0 & 41.4 & 41.9 & 39.2 & 57.7 & 45.3 & 44.6 & 43.8 \\
Qwen3.6-35B-A3B & 41.1 & 41.5 & 41.6 & 42.1 & 41.6 & 39.9 & 38.9 & 56.6 & 46.1 & 44.5 & 43.4 \\
Gemma-3-27B-it & 43.4 & 42.7 & 42.2 & 40.9 & 40.1 & 41.0 & 39.0 & 61.1 & 40.1 & 37.5 & 42.8 \\
Gemma-4-26B-A4B-it & 45.9 & 46.8 & 44.4 & 46.0 & 43.2 & 43.4 & 39.9 & 65.6 & 29.4 & 20.7 & 42.5 \\
Qwen3.5-9B & 40.1 & 39.2 & 39.1 & 39.4 & 39.5 & 39.3 & 37.6 & 55.2 & 39.7 & 37.9 & 40.7 \\
InternVL3.5-8B & 39.4 & 40.6 & 39.4 & 39.8 & 39.2 & 39.0 & 37.1 & 58.5 & 37.1 & 36.5 & 40.7 \\
Qwen3.5-35B-A3B & 40.4 & 38.9 & 38.1 & 38.3 & 38.4 & 37.9 & 36.8 & 54.9 & 39.7 & 37.0 & 40.0 \\
Molmo2-8B & 41.3 & 40.8 & 42.2 & 41.7 & 37.0 & 38.8 & 37.0 & 57.5 & 33.2 & 29.7 & 39.9 \\
Gemma-3-12B-it & 38.3 & 38.5 & 38.6 & 38.7 & 36.2 & 37.2 & 35.7 & 55.0 & 33.9 & 32.9 & 38.5 \\
MiniCPM-V-4.5 & 34.7 & 31.8 & 33.3 & 34.9 & 34.0 & 32.8 & 32.4 & 52.7 & 28.0 & 27.5 & 34.2 \\
Gemma-4-E4B-it & 28.0 & 29.9 & 27.0 & 26.4 & 26.7 & 27.8 & 24.8 & 47.6 & 16.9 & 15.3 & 27.0 \\
LLaVA-v1.6-Vicuna-13B & 16.5 & 16.1 & 16.2 & 16.2 & 16.3 & 16.2 & 15.0 & 31.4 & 2.9 & 2.8 & 14.9 \\
LLaVA-v1.6-Vicuna-7B & 15.2 & 15.3 & 15.0 & 14.5 & 14.7 & 14.9 & 14.1 & 28.2 & 3.8 & 3.0 & 13.9 \\
Table-LLaVA-v1.5-13B & 12.1 & 12.2 & 11.0 & 12.5 & 12.1 & 12.6 & 11.6 & 16.8 & 1.6 & 1.8 & 10.4 \\
Table-LLaVA-v1.5-7B & 9.2 & 9.1 & 7.5 & 10.1 & 9.2 & 9.4 & 8.8 & 13.8 & 0.9 & 1.1 & 7.9 \\
LLaVA-v1.5-7B & 6.4 & 6.2 & 6.2 & 7.0 & 6.9 & 6.7 & 6.4 & 10.4 & 0.5 & 0.4 & 5.7 \\
\bottomrule[1pt]
\end{tabular}
}
\caption{Model performance under different real-world scenarios, visual transformations, and vision-only settings.}
\label{tab:robustness_results}
\end{table*}

\paragraph{Structural Complexity Challenges.}
Complex structural layouts pose substantial challenges for current MLLMs, though the pattern diverges from simple intuition. For the frontier model GPT-5.4, accuracy follows the expected structural hierarchy, peaking on Text-Mixed (86.7\%) and Complex (81.7\%) tables and dropping sharply on Long (68.9\%) and Multi-table (61.3\%) scenarios. For weaker models, however, even Simple grids remain challenging (\eg Qwen2.5-VL-72B at 52.3\%), since our QA pairs demand intensive reasoning that a simple layout alone cannot offset. The Multi-table scenario proves uniquely difficult across the board, with most open-source models failing to exceed 50\%, suggesting that \ours requires spatial reasoning capabilities that remain difficult for current models, especially when context must be maintained across disjoint regions.

\paragraph{Reasoning Intensity and Table Skills.}
Performance varies considerably across reasoning difficulties mirroring the skill stratification defined in \S\ref{sec:3}. GPT-5.4 achieves 93.6\% on Easy tasks but attains just 47.0\% on Hard tasks requiring multi-hop aggregation. The gap widens further for open-source models, with even the strongest Qwen2.5-VL-72B collapsing from 90.2\% to 22.1\%. This uneven performance confirms that solving visual grounding does not automatically equate to solving complex tabular logic and highlights the need for future models to improve reasoning depth.

\paragraph{Proprietary vs. Open-source Models.}
Open-source foundation models remain notably behind proprietary models, with a substantial gap that persists across evaluations. Across all evaluated tasks, proprietary models consistently outperform open-source models with GPT-5.4 achieving 73.6\% overall accuracy. The top open-source model Qwen2.5-VL-72B achieves 55.5\% and trails GPT-5.4 by roughly 18 percentage points. However, this model significantly outperforms older architectures like LLaVA-v1.5-7B which fails to reach 7\% accuracy. This highlights continued progress in open-source development yet also points to substantial remaining challenges in achieving robust multimodal table reasoning.

\subsection{Evaluation on Real-World Scenario and Visual Transformation}
\label{sec:exp_robustness}
To assess the robustness of MLLMs against distribution shifts, we evaluate performance under diverse rendering styles and visual perturbations. The experiment results are presented in \autoref{tab:robustness_results}.

\paragraph{Scenario-Specific Analysis.}
Models exhibit remarkable consistency across different table scenarios. Contrary to the structural sensitivity observed in previous sections, performance remains surprisingly stable across \textit{Web}, \textit{LaTeX}, \textit{Excel}, and \textit{Customized} themes. For instance, GPT-5.4 maintains an accuracy around 72\% across all four styles, and the leading open-source model Qwen2.5-VL-72B shows similar stability with scores hovering around 55\%. This uniformity suggests that current foundation models have successfully generalized to diverse visual aesthetics and font variations.

\paragraph{Visual Transformation Sensitivity.}
Visual perturbations reveal distinct vulnerability patterns in model robustness. Counter-intuitively, the \textit{Missing} setting, which masks arbitrary cells while preserving answer-sufficient content, yields the highest scores across nearly all models—GPT-5.4 rises to 84.8\% and Qwen2.5-VL-72B reaches 71.3\%. This indicates that reducing visible clutter can actually concentrate model attention onto reasoning-critical cells. In contrast, the \textit{Partial} setting, which fragments the table into disjoint visual blocks, proves the most challenging overall, with GPT-5.4 dropping to 68.8\% and most open-source models falling below 50\%. \textit{Noise} and \textit{Structural Noise} perturbations cause moderate and comparable degradation, suggesting that current models tolerate low-level visual noise reasonably well but struggle to maintain spatial coherence once the table's structural continuity is broken.

\paragraph{Vision-Only Setting Analysis.}
The \textit{Vision-Only} setting reveals how models cope when questions and table content must be jointly parsed from a single image. Under the \textit{Screen Capture} mode, capable models largely retain their baseline competence—GPT-5.4 scores 69.4\% and Qwen2.5-VL-72B reaches 57.4\%, comparable to their performance with separately provided inputs. The \textit{Simulated Photo} mode, which overlays synthetic moiré patterns and perspective distortion on the rendered interface, induces further degradation: GPT-5.4 drops modestly to 67.3\%, while mid-sized models deteriorate more sharply (\eg Gemma-4-26B-A4B-it falls from 29.4\% to 20.7\%). A distinct failure mode emerges for legacy architectures such as the LLaVA family, whose accuracy collapses to near-zero under both vision-only modes, indicating an inability to extract questions and tabular content simultaneously from a unified visual input. Together, these results expose two separate bottlenecks: robustness to camera-style artifacts for capable models, and the more fundamental challenge of synergistic visual perception for weaker ones.

\begin{table}[t]
\centering
\setlength{\tabcolsep}{6pt}
\renewcommand\arraystretch{1.1}
\small
\begin{tabular}{l ccc}
\toprule[1pt]
\textbf{Model} & \textbf{DO} & \textbf{CoT} & \textbf{$\Delta$} \\
\midrule[0.7pt]
GPT-5.4             & 72.1 & 95.6 & +23.5 \\
GPT-5.4-mini        & 49.4 & 91.5 & +42.1 \\
Qwen3.5-27B         & 51.4 & \textbf{96.2} & \textbf{+44.8} \\
Gemma-4-31B-it      & 54.3 & 86.1 & +31.8 \\
Qwen3-VL-8B$^{\dagger}$ & 43.8 & 86.0 & +42.2 \\
\bottomrule[1pt]
\end{tabular}
\caption{Overall accuracy (\%) under direct-output (DO) and CoT prompting. $^{\dagger}$: Direct output uses Qwen3-VL-8B-Instruct; CoT uses Qwen3-VL-8B-Thinking.}
\label{tab:cot_vs_do}
\end{table}

\subsection{CoT vs. Direct-Output Prompting}
\label{sec:cot_vs_do}
Building on the DO-based analyses in \S\ref{sec:exp_complexity} and \S\ref{sec:exp_robustness}, we now quantify the landscape-compression effect noted in \S\ref{sec:exp_setup}. \autoref{tab:cot_vs_do} contrasts direct-output (DO) and chain-of-thought (CoT) prompting for five representative models. All models gain from explicit reasoning, but the gains are inversely tied to DO strength: GPT-5.4 rises only +23.5 points (72.1 to 95.6), while weaker DO performers close most of the gap (GPT-5.4-mini +42.1, Qwen3.5-27B +44.8, Gemma-4-31B-it +31.8). The effect is most striking at small scale: Qwen3-VL-8B jumps from 43.8 to 86.0 (+42.2), nearly erasing the gap to far larger models under CoT. The DO bottleneck on \ours thus lies less in visual perception, which \S\ref{sec:exp_robustness} shows to be handled well across rendering styles, than in packing multi-step tabular computation into a single forward pass. CoT also reshuffles the leaderboard: Qwen3.5-27B overtakes GPT-5.4 at the top, and the Gemma-Qwen ordering flips by over ten points. These contrasts position DO and CoT as complementary rather than redundant: CoT approaches the ceiling and reveals whether the required capabilities are present when reasoning is externalized, while DO stresses whether they can be internalized end-to-end and still leaves substantial headroom across all evaluated models.

\subsection{Qualitative Analysis}
\label{sec:qualitative_analysis}

We conduct a comprehensive qualitative audit to gain deeper insight into failure modes that aggregate metrics do not capture. Specifically, we randomly sample 100 examples from the error cases produced by GPT-5.4 and Qwen3.5-27B. We then manually analyze these examples to characterize the underlying failures and identify three primary failure modes:

\paragraph{Visual Brittleness.}
This case highlights incomplete visual recognition under noisy input as the primary failure mode. As shown in \autoref{fig:qua1}, GPT-5.4 under clean rendering correctly executes a Filter $\rightarrow$ Discriminate $\rightarrow$ Select pipeline and identifies \emph{Product-differentiated} ($\Delta\!=\!-17$\,pp vs.\ Scale-based's $-12$\,pp). Under simulated photo, however, degradation in the upper image region leads to partial question understanding: while the model correctly extracts all relevant numerical values, it fails to recognize the superlative qualifier \emph{largest}. This selective omission causes the model to operate on incomplete information—after identifying sectors satisfying the base condition, its internal objective is effectively reduced from \emph{Filter-and-Argmax} to \emph{Filter-and-Verify}, resulting in the output \{Scale-based, Product-differentiated\}. The error thus reflects localized visual brittleness: missing fine-grained textual cues leads to answers that are internally consistent with perceived inputs, yet incomplete with respect to the full query, without any calibration signal to indicate the mismatch.

\paragraph{Spatial Alignment Failures.}
Structural complexity places substantial demands on spatial reasoning, especially in dense tables with multi-level headers. As shown in \autoref{fig:qua2}, a representative \textit{cross-row argmin} error arises at the filtering stage. Under clean rendering, GPT-5.4 correctly compares \emph{Men}\,=\,15.2 and \emph{Women}\,=\,11.8, selects \emph{Women}, and returns 22.9. Under simulated photo conditions, however, the reasoning trace remains logically consistent while perception fails: \(11.8\) is misread as \(15.8\) due to a \(1\!\to\!5\) substitution influenced by the nearby \(15.2\). This misperception flips the comparison, leading the model to select \emph{Men} and retrieve 16.3\,pp, a valid but incorrect cell. Notably, row alignment remains intact; the failure originates from fine-grained digit confusion, which propagates to the decision stage while remaining undetected in the CoT. This highlights a key limitation: coarse structural grounding is insufficient, as models struggle to distinguish visually similar digits within correctly localized cells without stronger sub-cell disambiguation.

\begin{figure}[t]
  \centering
  \includegraphics[width=0.8\linewidth]{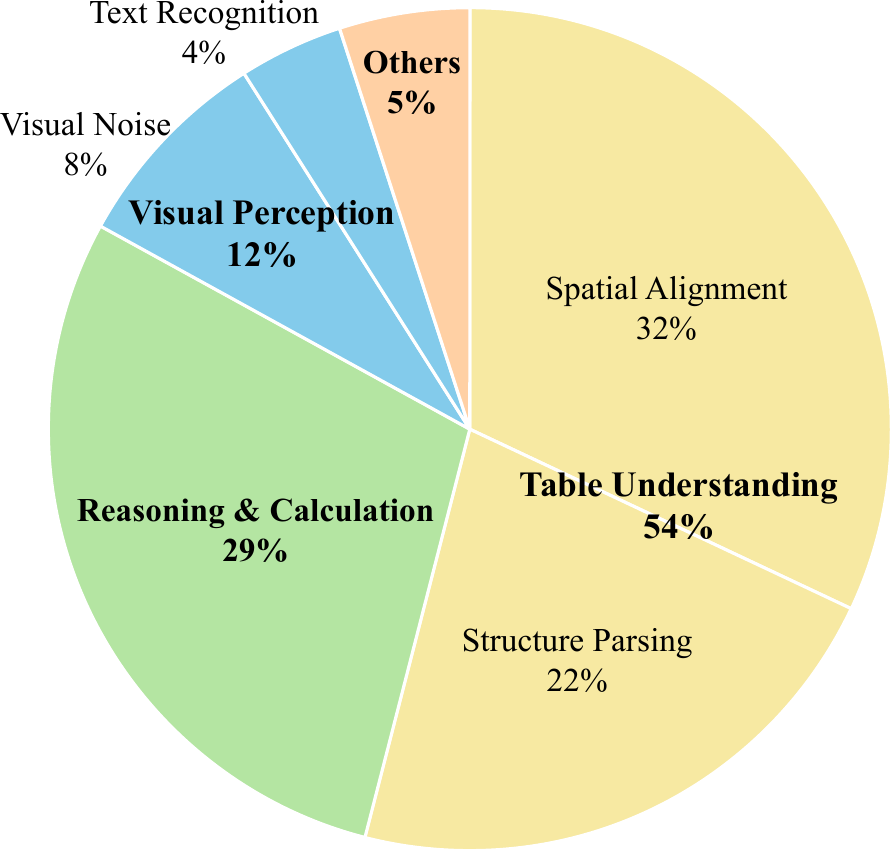}
  \caption{Error distribution analysis on \ours. The majority of failures stem from Table Understanding (54\%), followed by Reasoning \& Calculation (29\%) and Visual Perception (12\%).}
  \label{fig:error_analysis}
\end{figure}
\paragraph{Error Distribution.}
To pinpoint failure causes, we analyzed the error distribution, as shown in \autoref{fig:error_analysis}. \textit{Table Understanding} dominates the error distribution (54\%), significantly surpassing \textit{Reasoning \& Calculation} (29\%) and \textit{Visual Perception} (12\%). Within table understanding, \textit{Spatial Alignment} (32\%) and \textit{Structure Parsing} (22\%) are the primary bottlenecks. Notably, errors stemming solely from text recognition are rare (4\%), confirming that the challenge lies in maintaining spatial structure rather than basic perception.

\section{Conclusion}
We present \ours, a comprehensive benchmark designed to rigorously evaluate the multimodal table reasoning capabilities of foundation models under visual and structural complexity. Each instance in \ours is expanded into distinct visual variants through a rigorous multi-style rendering pipeline and audited by human experts. Our evaluation of \nmodel frontier foundation models reveals that even the most advanced proprietary models degrade sharply as structural and reasoning complexity grow, remain brittle under realistic visual conditions, and show pronounced open-source to proprietary gaps under direct-output prompting. These findings expose critical limits in anchoring structured reasoning in visual space, providing valuable insights for advancing the robust and structure-aware reasoning abilities of future foundation models.
\section*{Limitations}
This work focuses on evaluating frontier models for robust multimodal table understanding. We do not explore or propose new training methods to directly mitigate the structural alignment failures identified in our tasks. Future research could build upon our analyses to investigate effective visual instruction tuning strategies that enhance spatial grounding in complex document layouts.
Moreover, beyond the tabular data covered in our benchmark, real-world documents often involve interleaved modalities such as charts and natural images. We therefore encourage future work to extend our study by incorporating a broader range of visual document elements and heterogeneous information sources.

\bibliography{anthology,custom}

\appendix
\clearpage
\begin{table*}[!t]
\centering
\small
\setlength{\tabcolsep}{10pt}
\renewcommand{\arraystretch}{1.35}
\rowcolors{2}{black!3}{white}
\begin{tabularx}{\textwidth}{c l l X}
\toprule
\textbf{ID} & \textbf{Education} & \textbf{Major} & \textbf{Assigned Task} \\
\midrule
1  & PhD Student (Yr 3)    & Computer Science        & Attribute Labeling, QA Enhancement, Visual Rendering Audit, Quality Audit \\
2  & Master (Yr 2)         & Computer Science        & Attribute Labeling, QA Enhancement \\
3  & PhD Student (Yr 2)    & Computer Science        & Attribute Labeling, QA Enhancement, Visual Rendering Audit \\
4  & PhD Student (Yr 4)  & Computer Science        & Attribute Labeling, QA Enhancement, Quality Audit \\
5  & PhD Student (Yr 2)    & Computer Science        & Attribute Labeling, QA Enhancement, Visual Rendering Audit \\
6  & Master (Yr 2)         & Computer Science        & Attribute Labeling, QA Enhancement \\
7  & PhD Student (Yr 1)    & Data Science            & Attribute Labeling, QA Enhancement \\
8  & Master (Yr 2)         & Data Science            & Attribute Labeling, QA Enhancement, Visual Rendering Audit \\
9  & PhD Student (Yr 3)    & Statistics              & Attribute Labeling, QA Enhancement, Quality Audit \\
10 & PhD Student (Yr 2)    & Information Systems     & Attribute Labeling, QA Enhancement \\
11 & Master (Yr 2)         & Economics               & Attribute Labeling, QA Enhancement \\
12 & PhD Student (Yr 1)    & Mathematics             & Attribute Labeling, QA Enhancement \\
\bottomrule
\end{tabularx}
\caption{%
    Profile of the 12 expert annotators recruited for \ours. All annotators participated in \textit{Attribute Labeling} (table layout, information richness, skill scores, and reasoning steps) and \textit{QA Enhancement} (question rephrasing and deep synthesis) over the base set. Four annotators additionally conducted the \textit{Visual Rendering Audit}, verifying answerability under structural and visual perturbations (\eg \textit{Missing} transformations). Three senior annotators further performed the final \textit{Quality Audit}, reviewing the full benchmark for consistency, answerability, and difficulty calibration.
}
\label{tab:experts}
\end{table*}
\begin{table*}[!t]
\centering
\small
\setlength{\tabcolsep}{6pt}
\renewcommand{\arraystretch}{1.35}
\begin{tabularx}{\textwidth}{l X r r r}
\toprule
\textbf{Phase} & \textbf{Stage} & \textbf{Initial} & \textbf{Removed / Revised} & \textbf{Rate} \\
\midrule
  & Taxonomic Categorization and Multidimensional Filtering       & 100{,}000 & 97{,}000 removed     & 97.00\% \\
  & QA Enhancement -- Question Rephrasing                    & 3{,}000   & 2{,}264 rephrased    & 75.47\% \\
\multirow{-3}{*}{Base Set}
  & QA Enhancement -- Deep Synthesis (regeneration)          & 3{,}000   & 736 regenerated      & 24.53\% \\
\midrule
  & Rendering Failure Handling                              & 3{,}000   & 14 replaced          & 0.47\%  \\
  & Answerability Verification \& Calibration               & 30{,}000  & 912 regenerated      & 3.04\%  \\
\multirow{-3}{*}{Visual Rendering}
  & Cross Validation (10\% of base samples, double-blind)   & 300       & 6 corrected          & 2.00\%  \\
\bottomrule
\end{tabularx}
\caption{Rejection and revision statistics at each stage of the \ours construction pipeline. The Base Set phase filters 100{,}000 attribute-labeled candidates down to 3{,}000 samples with high structural and reasoning diversity, followed by QA Enhancement in which every sample is either rephrased or rewritten from scratch. The Visual Rendering phase produces 30{,}000 images across 10 categories and applies three successive rounds of verification, yielding a final benchmark with fully verified answerability and minimal residual annotation disagreement.
}
\label{tab:quality_control}
\end{table*}
\section{Annotator Information} \label{app:annotators}
To ensure the high quality of \ours, we recruited 12 expert annotators. All annotators are graduate students (Master's or PhD) in quantitative or technical fields such as computer science, data science, statistics, and mathematics. Prior to the main annotation, each annotator completed a standardized 1-hour tutorial on our protocols and a 30-sample pilot annotation, which was reviewed and approved by the project leads before proceeding. \autoref{tab:experts} provides detailed profiles of the annotation team.

For quality assurance, every sample was reviewed by at least one annotator during the main annotation pass, and 10\% of samples were independently cross-validated by a second annotator, with disagreements resolved through consensus discussion. Finally, all samples were passed through automated checks for answer uniqueness and format consistency before final approval.

\section{Quality Control Statistics} \label{app:quality_control}
To provide full transparency regarding the rigor of our data construction pipeline, we report the rejection and revision statistics at every stage, based on our original annotation and audit logs. The pipeline proceeds in two phases---Base Set Construction and Visual Rendering Pipeline---each subject to multiple rounds of filtering, verification, and cross-validation. \autoref{tab:quality_control} summarizes the quantitative outcomes of each stage.

\paragraph{Base Set Construction.}
We first aggregated raw records from the 14 public datasets and performed AI-assisted preliminary cleaning to remove low-quality samples while maintaining a balanced source distribution, retaining approximately 100{,}000 candidate samples. All 100{,}000 samples then underwent AI-assisted \textit{Attribute Labeling} with manual verification as described in \S\ref{sec:3.1}, covering table layout attributes and dimensions, \textit{information richness} scores, \textit{skill scores} across the four reasoning dimensions, and the number of \textit{reasoning steps}. Based on these labeled attributes, we applied \textit{Taxonomic Categorization and Multidimensional Filtering} to select 3{,}000 samples that jointly exhibit strong structural diversity and high reasoning complexity, corresponding to a 97.0\% rejection rate. For the retained 3{,}000 samples, \textit{QA Enhancement} was performed under the dual-track strategy described in \S\ref{sec:3.1}: 2{,}264 samples were rephrased to increase linguistic diversity and reduce potential contamination, while the remaining 736 samples were authored from scratch as new multi-step reasoning questions grounded in the table and textual context.

\paragraph{Visual Rendering Pipeline.}
The 3{,}000 textual samples were rendered into the 10 visual settings described in \S\ref{sec:3.2}, yielding 30{,}000 images in total. Three rounds of quality control were applied. First, in \textit{Rendering Failure Handling}, 14 samples that failed rendering due to aspect ratio or context-length constraints were replaced to ensure completeness. Second, in \textit{Answerability Verification and Parameter Calibration}, we repeatedly sampled 100 images per setting to assess post-render answerability and iteratively tuned the rendering parameters until the pipeline reached 97.2\% answerability; after fixing the parameters, we manually verified answerability for all 30{,}000 rendered images and regenerated 912 images (3.04\%) whose content had been compromised by the visual transformations. Finally, as part of the \emph{human-in-the-loop} audit introduced in \S\ref{sec:3.3}, 10\% of the base samples (300 samples together with all their rendered variants) were subjected to double-blind cross-validation by two independent annotators; only 6 samples required minor corrections, corresponding to a 2.0\% disagreement rate and indicating that the earlier audit stages had already resolved the vast majority of quality issues.

Overall, these statistics demonstrate that \ours is the product of a multi-stage, human-in-the-loop pipeline in which every sample is filtered, enhanced, rendered, and independently verified.


\section{Experiment Setup}
\subsection{Configuration of Evaluated Models} \label{app:model_info}
\autoref{tab:model_configuration} lists the full inference configuration of each evaluated model.
Rather than imposing a single uniform decoding setting, we follow the sampling parameters (temperature, top-$p$, top-$k$, and repetition penalty) recommended by each model's official release, since recommended defaults vary substantially across model families; models without an explicit sampling recommendation are evaluated with greedy decoding.
The maximum number of generated tokens is set to 8{,}192 for open-source models, except for the LLaVA family, whose native 4{,}096-token context window restricts generation to 2{,}048 tokens.
For closed-source models accessed through official API endpoints, we use the provider's default sampling configuration and cap the completion length at 16{,}000 tokens.
All open-source models are served locally via vLLM with tensor-parallel sizes ranging from 1 to 8 depending on model scale.
Our experiments are conducted on a workstation equipped with multiple NVIDIA H20 GPUs.

\subsection{Prompts}\label{app:prompt}
We release the full set of prompts used in our evaluation. As discussed in \S\ref{sec:cot_vs_do}, all main results in \S\ref{sec:exp_complexity} and \S\ref{sec:exp_robustness} adopt the \textit{direct-output} template (\autoref{fig:prompt_do}), while the CoT variant (\autoref{fig:prompt_cot}) is used in the CoT-vs-direct-output analysis. Each template comes with two modalities: a \textit{general} version for the standard image+text setting, and a \textit{vision-only} version for the \textit{Screen Capture} and \textit{Simulated Photo} configurations where the question is rendered into the image. In addition, the LLM-as-Judge prompt (\autoref{fig:prompt_judge}) is applied as a secondary verification layer on predictions that fail the Exact Match criterion.

\clearpage
\begin{table*}[!t]
\centering
\footnotesize
\resizebox{\textwidth}{!}{%
\begin{tabular}{lllccc}
\toprule
\textbf{Organization} & \textbf{Model} & \textbf{Version} & \textbf{\makecell{Serving\\Backend}} & \textbf{Temp.} & \textbf{\makecell{Max\\Tokens}} \\
\midrule
\multicolumn{6}{c}{\emph{\textbf{Proprietary Models}}} \\
\midrule
\multirow{2}{*}{OpenAI}
& GPT-5.4      & \texttt{gpt-5.4-0305-global}      & API & -- & 16{,}000 \\
& GPT-5.4-mini & \texttt{gpt-5.4-mini-0317-global} & API & -- & 16{,}000 \\
\midrule
\multicolumn{6}{c}{\emph{\textbf{Open-Source Foundation Models}}} \\
\midrule
\multirow{10}{*}{Qwen}
& Qwen3.6-35B-A3B            & \texttt{Qwen/Qwen3.6-35B-A3B}            & vLLM & 1.0 & 8{,}192 \\
& Qwen3.5-122B-A10B          & \texttt{Qwen/Qwen3.5-122B-A10B}          & vLLM & 1.0 & 8{,}192 \\
& Qwen3.5-35B-A3B            & \texttt{Qwen/Qwen3.5-35B-A3B}            & vLLM & 1.0 & 8{,}192 \\
& Qwen3.5-27B                & \texttt{Qwen/Qwen3.5-27B}                & vLLM & 1.0 & 8{,}192 \\
& Qwen3.5-9B                 & \texttt{Qwen/Qwen3.5-9B}                 & vLLM & 1.0 & 8{,}192 \\
& Qwen3-VL-30B-A3B           & \texttt{Qwen/Qwen3-VL-30B-A3B-Instruct}  & vLLM & 0.7 & 8{,}192 \\
& Qwen3-VL-8B                & \texttt{Qwen/Qwen3-VL-8B-Instruct}       & vLLM & 0.7 & 8{,}192 \\
& Qwen2.5-VL-72B             & \texttt{Qwen/Qwen2.5-VL-72B-Instruct}    & vLLM & 0.0 & 8{,}192 \\
& Qwen2.5-VL-32B             & \texttt{Qwen/Qwen2.5-VL-32B-Instruct}    & vLLM & 0.0 & 8{,}192 \\
& Qwen2.5-VL-7B              & \texttt{Qwen/Qwen2.5-VL-7B-Instruct}     & vLLM & 0.0 & 8{,}192 \\
\noalign{\vskip 0.5ex}\hdashline\noalign{\vskip 0.5ex}
\multirow{3}{*}{InternVL}
& InternVL3.5-30B-A3B & \texttt{OpenGVLab/InternVL3\_5-30B-A3B} & vLLM & 0.0 & 8{,}192 \\
& InternVL3.5-14B     & \texttt{OpenGVLab/InternVL3\_5-14B}     & vLLM & 0.0 & 8{,}192 \\
& InternVL3.5-8B      & \texttt{OpenGVLab/InternVL3\_5-8B}      & vLLM & 0.0 & 8{,}192 \\
\noalign{\vskip 0.5ex}\hdashline\noalign{\vskip 0.5ex}
\multirow{2}{*}{Meta}
& Llama-4-Maverick & \texttt{meta-llama/Llama-4-Maverick-17B-128E-Instruct-FP8} & vLLM & 0.6 & 8{,}192 \\
& Llama-4-Scout    & \texttt{meta-llama/Llama-4-Scout-17B-16E-Instruct}         & vLLM & 0.6 & 8{,}192 \\
\noalign{\vskip 0.5ex}\hdashline\noalign{\vskip 0.5ex}
\multirow{5}{*}{Google}
& Gemma-4-31B-it     & \texttt{google/gemma-4-31B-it}     & vLLM & 1.0 & 8{,}192 \\
& Gemma-4-26B-A4B-it & \texttt{google/gemma-4-26B-A4B-it} & vLLM & 1.0 & 8{,}192 \\
& Gemma-4-E4B-it     & \texttt{google/gemma-4-E4B-it}     & vLLM & 1.0 & 8{,}192 \\
& Gemma-3-27B-it     & \texttt{google/gemma-3-27b-it}     & vLLM & 1.0 & 8{,}192 \\
& Gemma-3-12B-it     & \texttt{google/gemma-3-12b-it}     & vLLM & 1.0 & 8{,}192 \\
\noalign{\vskip 0.5ex}\hdashline\noalign{\vskip 0.5ex}
\multirow{1}{*}{Allen AI}
& Molmo2-8B & \texttt{allenai/Molmo2-8B} & vLLM & 0.0 & 8{,}192 \\
\noalign{\vskip 0.5ex}\hdashline\noalign{\vskip 0.5ex}
\multirow{1}{*}{OpenBMB}
& MiniCPM-V-4.5 & \texttt{openbmb/MiniCPM-V-4\_5} & vLLM & 0.6 & 8{,}192 \\
\noalign{\vskip 0.5ex}\hdashline\noalign{\vskip 0.5ex}
\multirow{3}{*}{LLaVA}
& LLaVA-v1.6-Vicuna-13B & \texttt{llava-hf/llava-v1.6-vicuna-13b-hf} & vLLM & 0.0 & 2{,}048 \\
& LLaVA-v1.6-Vicuna-7B  & \texttt{llava-hf/llava-v1.6-vicuna-7b-hf}  & vLLM & 0.0 & 2{,}048 \\
& LLaVA-v1.5-7B         & \texttt{llava-hf/llava-1.5-7b-hf}          & vLLM & 0.0 & 2{,}048 \\
\noalign{\vskip 0.5ex}\hdashline\noalign{\vskip 0.5ex}
\multirow{2}{*}{Table-LLaVA}
& Table-LLaVA-v1.5-13B & \texttt{SpursgoZmy/table-llava-v1.5-13b-hf} & vLLM & 0.0 & 2{,}048 \\
& Table-LLaVA-v1.5-7B  & \texttt{SpursgoZmy/table-llava-v1.5-7b-hf}  & vLLM & 0.0 & 2{,}048 \\
\bottomrule
\end{tabular}
}
\caption{
Details of the multimodal foundation models evaluated in \ours\ and their inference configurations.
The \textbf{Serving Backend} indicates how the model was accessed in our experiments (API for cloud endpoints; vLLM for local deployment on NVIDIA H20 GPUs).
\textbf{Temperature} follows each model's official recommendation; ``--'' indicates that we use the provider's default sampling configuration for API models.
Other sampling parameters (top-$p$, top-$k$, and repetition penalty, when applicable) also follow each model's official recommendation.
\textbf{Max Tokens} is the upper bound on generated tokens per response.
}
\label{tab:model_configuration}
\end{table*}

\begin{figure*}[!t]
\centering
\begin{minipage}{0.98\textwidth}
\begin{tcolorbox}[
  breakable, enhanced,
  colback=white, colframe=black!40,
  boxrule=0.5pt, arc=2pt,
  left=6pt, right=6pt, top=4pt, bottom=5pt,
  title={Direct-Output Evaluation Prompt},
  colbacktitle=black!15, coltitle=black,
  fonttitle=\normalsize\bfseries, fontupper=\small,
]

\textbf{\textit{(a) General --- Image+Text Mode}}
\vspace{2pt}\hrule height 0.3pt\vspace{3pt}

\begin{tcolorbox}[
  colback=blue!4, colframe=blue!35, boxrule=0.4pt, arc=1.5pt,
  left=6pt, right=6pt, top=3pt, bottom=3pt, fontupper=\small,
]
\textcolor{blue!55!black}{\scriptsize\textbf{\textsf{SYSTEM}}}\\[1pt]
You are a table understanding expert. Answer the question based on the given table image. Provide only the final answer, no explanation.
\end{tcolorbox}

\vspace{-3pt}

\begin{tcolorbox}[
  colback=black!3, colframe=black!40, boxrule=0.4pt, arc=1.5pt,
  left=6pt, right=6pt, top=3pt, bottom=3pt, fontupper=\small,
]
\textcolor{black!65}{\scriptsize\textbf{\textsf{USER}}}\\[1pt]
\texttt{[Table Image]}\\[2pt]
Answer the question based on the table shown in the image.\\[2pt]
\textbf{Question:} \texttt{\{question\}}\\[2pt]
Answer directly with the answer only.
\end{tcolorbox}

\vspace{6pt}
\textbf{\textit{(b) Vision-Only Mode}} \textnormal{\footnotesize (used for \textit{Screen Capture} and \textit{Simulated Photo})}
\vspace{2pt}\hrule height 0.3pt\vspace{3pt}

\begin{tcolorbox}[
  colback=blue!4, colframe=blue!35, boxrule=0.4pt, arc=1.5pt,
  left=6pt, right=6pt, top=3pt, bottom=3pt, fontupper=\small,
]
\textcolor{blue!55!black}{\scriptsize\textbf{\textsf{SYSTEM}}}\\[1pt]
You are a table understanding expert. The image contains both a question and a table. Read the question from the image and answer it based on the table. Provide only the final answer, no explanation.
\end{tcolorbox}

\vspace{-3pt}

\begin{tcolorbox}[
  colback=black!3, colframe=black!40, boxrule=0.4pt, arc=1.5pt,
  left=6pt, right=6pt, top=3pt, bottom=3pt, fontupper=\small,
]
\textcolor{black!65}{\scriptsize\textbf{\textsf{USER}}}\\[1pt]
\texttt{[Image containing both Question and Table]}\\[2pt]
The question is shown in the image above. Answer it based on the table in the image.\\[2pt]
Answer directly with the answer only.
\end{tcolorbox}

\end{tcolorbox}
\end{minipage}
\caption{Direct-output evaluation prompt. (a)~The general template supplies the table image and the textual question as separate inputs. (b)~The vision-only template is used when the question is rendered into the image together with the table, requiring the model to jointly parse both from a single visual input.}
\label{fig:prompt_do}
\end{figure*}

\begin{figure*}[!t]
\centering
\begin{minipage}{0.98\textwidth}
\begin{tcolorbox}[
  breakable, enhanced,
  colback=white, colframe=black!40,
  boxrule=0.5pt, arc=2pt,
  left=6pt, right=6pt, top=4pt, bottom=5pt,
  title={Chain-of-Thought Evaluation Prompt},
  colbacktitle=black!15, coltitle=black,
  fonttitle=\normalsize\bfseries, fontupper=\small,
]

\textbf{\textit{(a) General --- Image+Text Mode}}
\vspace{2pt}\hrule height 0.3pt\vspace{3pt}

\begin{tcolorbox}[
  colback=blue!4, colframe=blue!35, boxrule=0.4pt, arc=1.5pt,
  left=6pt, right=6pt, top=3pt, bottom=3pt, fontupper=\small,
]
\textcolor{blue!55!black}{\scriptsize\textbf{\textsf{SYSTEM}}}\\[1pt]
You are a table understanding expert. Answer the question based on the given table image.
\end{tcolorbox}

\vspace{-3pt}

\begin{tcolorbox}[
  colback=black!3, colframe=black!40, boxrule=0.4pt, arc=1.5pt,
  left=6pt, right=6pt, top=3pt, bottom=3pt, fontupper=\small,
]
\textcolor{black!65}{\scriptsize\textbf{\textsf{USER}}}\\[1pt]
\texttt{[Table Image]}\\[2pt]
Answer the question based on the table shown in the image.\\[2pt]
\textbf{Question:} \texttt{\{question\}}\\[2pt]
Think step by step, then give your final answer in the format: \texttt{<answer> ... </answer>}
\end{tcolorbox}

\vspace{6pt}
\textbf{\textit{(b) Vision-Only Mode}} \textnormal{\footnotesize (used for \textit{Screen Capture} and \textit{Simulated Photo})}
\vspace{2pt}\hrule height 0.3pt\vspace{3pt}

\begin{tcolorbox}[
  colback=blue!4, colframe=blue!35, boxrule=0.4pt, arc=1.5pt,
  left=6pt, right=6pt, top=3pt, bottom=3pt, fontupper=\small,
]
\textcolor{blue!55!black}{\scriptsize\textbf{\textsf{SYSTEM}}}\\[1pt]
You are a table understanding expert. The image contains both a question and a table. Read the question from the image and answer it based on the table.
\end{tcolorbox}

\vspace{-3pt}

\begin{tcolorbox}[
  colback=black!3, colframe=black!40, boxrule=0.4pt, arc=1.5pt,
  left=6pt, right=6pt, top=3pt, bottom=3pt, fontupper=\small,
]
\textcolor{black!65}{\scriptsize\textbf{\textsf{USER}}}\\[1pt]
\texttt{[Image containing both Question and Table]}\\[2pt]
The question is shown in the image above. Answer it based on the table in the image.\\[2pt]
Think step by step, then give your final answer in the format: \texttt{<answer> ... </answer>}
\end{tcolorbox}

\end{tcolorbox}
\end{minipage}
\caption{Chain-of-Thought evaluation prompt. (a)~The general template elicits step-by-step reasoning before the final tagged answer. (b)~The vision-only template applies the same CoT protocol when the question and table are provided together as a single image. For models supporting native thinking mode (\eg Qwen3-VL, Qwen3.5), we additionally enable \texttt{enable\_thinking=True} at inference.}
\label{fig:prompt_cot}
\end{figure*}

\begin{figure*}[!t]
\centering
\begin{minipage}{0.98\textwidth}
\begin{tcolorbox}[
  colback=white, colframe=black!40,
  boxrule=0.5pt, arc=2pt,
  left=6pt, right=6pt, top=4pt, bottom=5pt,
  title={LLM-as-Judge Evaluation Prompt},
  colbacktitle=black!15, coltitle=black,
  fonttitle=\normalsize\bfseries, fontupper=\small,
]

\textnormal{\footnotesize Applied to predictions that fail Exact Match, using GPT-5-mini as the judge model.}

\vspace{3pt}\hrule height 0.3pt\vspace{5pt}

{\color{black!65}\scriptsize\textbf{\textsf{USER}}}

\vspace{2pt}
You are evaluating a table question answering system.

\vspace{3pt}
\textbf{Question:} \texttt{\{question\}}\\
\textbf{Ground truth answer:} \texttt{\{answer\}}\\
\textbf{Predicted answer:} \texttt{\{predicted\}}

\vspace{3pt}
Mark as CORRECT (\texttt{`yes'}) if any of the following apply:
\begin{itemize}[leftmargin=1.4em, topsep=1pt, itemsep=1pt, parsep=0pt, label=\textbullet]
    \item The prediction matches the gold answer exactly or semantically (case, abbreviation, minor formatting differences are OK).
    \item The prediction contains the gold answer and only adds valid supplementary information such as units or clarifying context (\eg gold=\texttt{`713'}, pred=\texttt{`713 thousand shares'} is OK).
\end{itemize}

\vspace{2pt}
Mark as INCORRECT (\texttt{`no'}) otherwise.

\vspace{3pt}
\textit{Numeric substring trap:} gold=\texttt{`2'} $\neq$ pred=\texttt{`20'}, gold=\texttt{`13'} $\neq$ pred=\texttt{`13.5'}.

\vspace{3pt}
Reply with ONLY \texttt{`yes'} or \texttt{`no'}.

\end{tcolorbox}
\end{minipage}
\caption{LLM-as-Judge evaluation prompt. Applied as a secondary verification layer on predictions that fail the Exact Match criterion, tolerating formatting variations while rejecting numeric substring collisions. We use GPT-5-mini as the judge throughout all experiments.}
\label{fig:prompt_judge}
\end{figure*}

\clearpage
\section{Qualitative Analysis}
\begin{figure*}[!t]
    \centering
    \includegraphics[width=\textwidth]{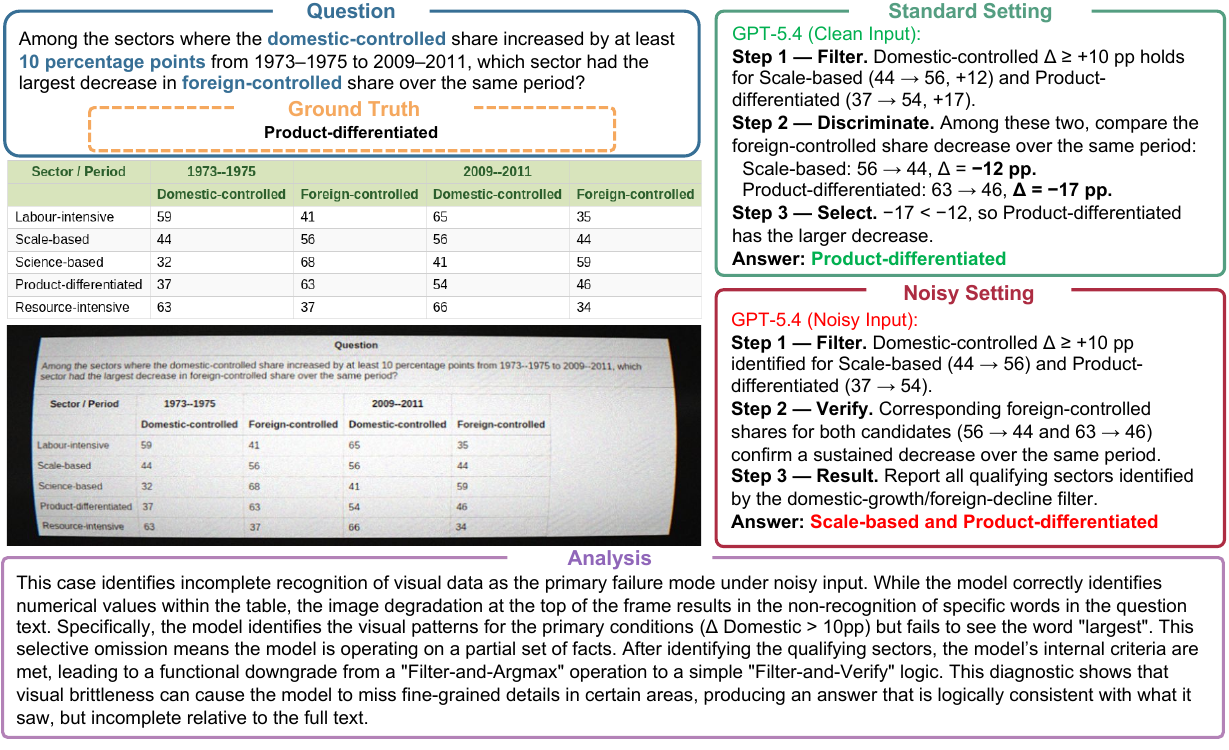}
    \caption{Qualitative analysis of visual brittleness in \ours.}
    \label{fig:qua1}
\end{figure*}

\begin{figure*}[!t]
    \centering
    \includegraphics[width=\textwidth]{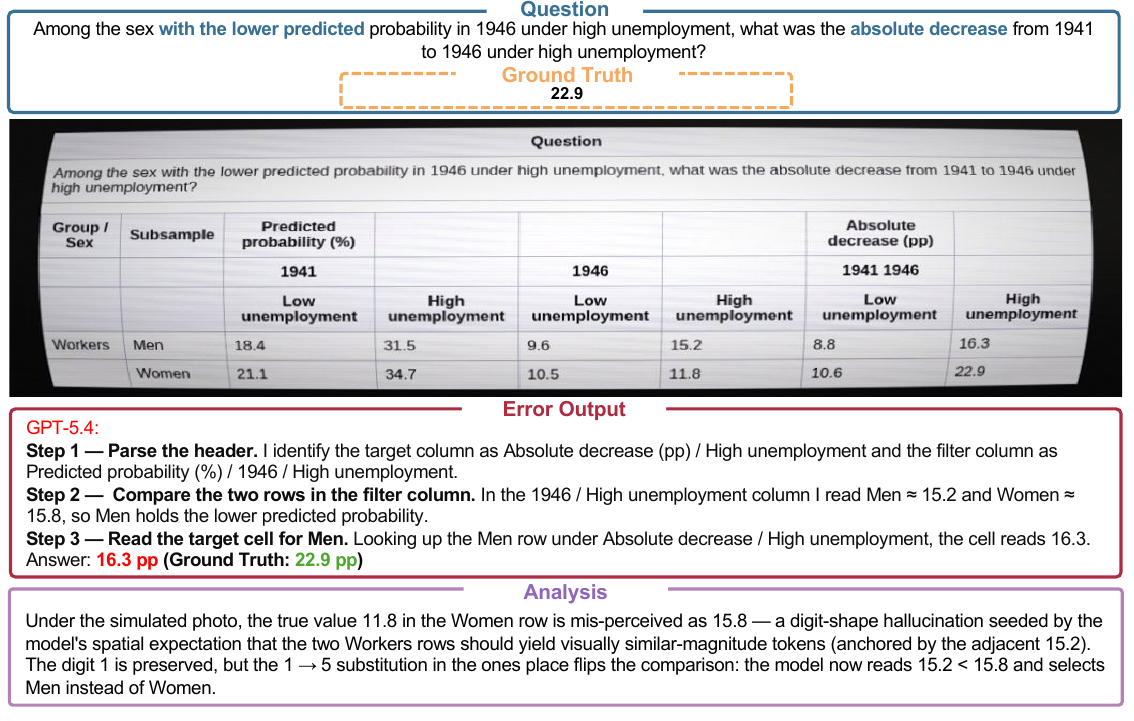}
    \caption{Qualitative analysis of spatial alignment failures in \ours.}
    \label{fig:qua2}
\end{figure*}

To gain deeper insights into the failure modes of multimodal foundation models beyond aggregate metrics, we present detailed case studies in \ours. We examine the impact of visual perturbations on reasoning consistency in \autoref{fig:qua1} and the challenges of spatial grounding in complex layouts in \autoref{fig:qua2}.

\section{Visual Examples}
\begin{figure*}[!t]
    \centering
    \includegraphics[width=\textwidth]{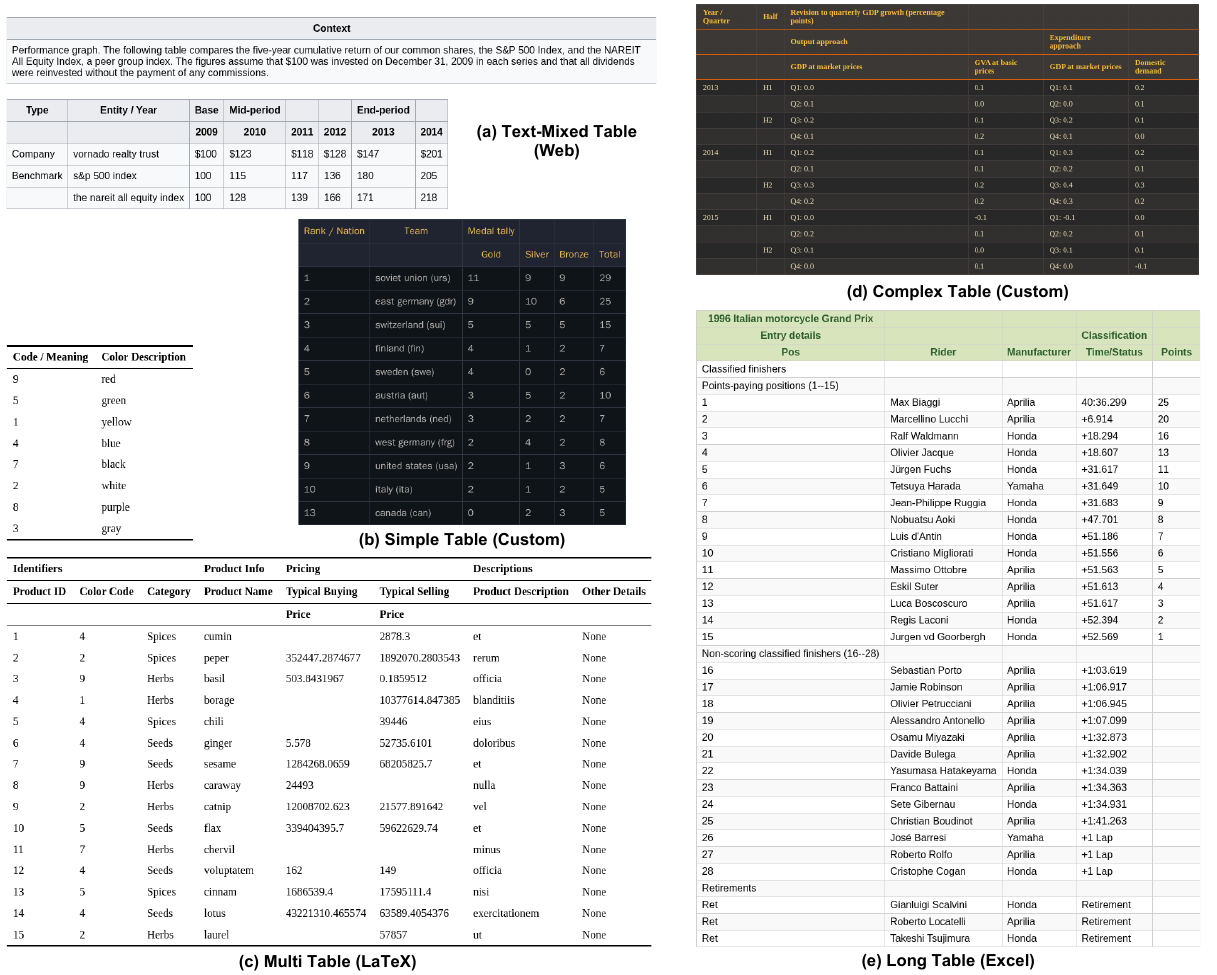}
    \caption{Visual examples for different table structures.}
    \label{fig:example1}
\end{figure*}

To demonstrate the structural diversity and complexity of the \ours benchmark, we present representative samples across different categories. \autoref{fig:example1} displays examples of the five table structural types defined in our taxonomy, ranging from standard grids to hierarchical and multi-table layouts, verifying the breadth of challenges covered in our dataset.

\end{document}